\documentclass{ifacconf}

\usepackage{graphicx}      
\usepackage{natbib}        
\usepackage{graphics,color}
\usepackage[outdir=./]{epstopdf}
\usepackage{amsmath}
\usepackage{amssymb}
\usepackage{mathrsfs}
\usepackage{subfigure}
\usepackage[font=small,skip=5pt]{caption}
\usepackage{url}
\usepackage{float}
\usepackage{bbm}
\usepackage{booktabs}
\usepackage{array}
\usepackage[table]{xcolor}

\usepackage{enumitem}
\usepackage{optidef} 
\usepackage{mdframed}

\newtheorem{remark}{Remark}

\newtheorem{example}{Example}

\newcommand{\real}{\mathbb{R}}

\newcommand*{\QEDW}{\hfill\ensuremath{\square}}

\DeclareMathOperator*{\argmin}{arg\,min}

\newcommand\oprocendsymbol{\hbox{$\square$}}
\newcommand\oprocend{\relax\ifmmode\else\unskip\hfill\fi\oprocendsymbol}

\usepackage{tabstackengine}
\setstackEOL{\cr}

\begin{document}
\begin{frontmatter}

\title{Alternating Minimization for Time-Shifted Synergy Extraction in Human Hand Coordination}



\author{Trevor Stepp,} 
\author{Parthan Olikkal,}
\author{Ramana Vinjamuri, and} 
\author{Rajasekhar Anguluri} 


\address{Department of Computer Science and
Electrical Engineering, University of Maryland, Baltimore County, MD 21229 USA \\ e-mails: \{trevors3, polikka1, rvinjam1, rajangul\}@umbc.edu}

\begin{abstract}                
Identifying motor synergies -- coordinated hand joint patterns activated at
task-dependent time shifts -- from kinematic data is central to motor
control and robotics. Existing two-stage methods first extract candidate
waveforms (via SVD) and then select shifted templates using sparse
optimization, requiring at least two datasets and complicating data
collection.

\medskip  
We introduce an optimization-based framework that jointly learns a small
set of synergies and their sparse activation coefficients. The formulation
enforces group sparsity for synergy selection and element-wise sparsity for
activation timing. We develop an alternating minimization method in which
coefficient updates decouple across tasks and synergy updates reduce to regularized least-squares problems. Our approach requires only a
single data set, and simulations show accurate velocity reconstruction with
compact, interpretable synergies. 

\medskip 

{\it Keywords:} Human hand coordination, Convolution-mixture model, Dimensionality reduction, Group LASSO, Alternating minimization.
\end{abstract}

\end{frontmatter}

\smallskip 
\section{Introduction}

\vspace{-2.0mm}

Our hand movements exhibit remarkable flexibility despite the
high-dimensional nature of the musculoskeletal system. Coordinating
dozens of joints and muscles in real time presents a challenging 
control problem. Yet, we perform grasping and manipulation tasks 
with ease, suggesting that the central nervous system exploits 
low-dimensional structure in hand motion. This idea is captured 
through the notion of \emph{motor synergies}: coordinated patterns of hand joint movements that serve as building blocks for 
executing a wide range of tasks (see \cite{bernstein1967coordination}). 

A key question in motor control and robotics is therefore how to 
identify such synergies directly from measured kinematic data. 
Dimensionality reduction plays a key role here: representing hand motion 
using only a few synergies not only yields compact models but also provides 
interpretability relevant to neuroscience, prosthetics, human--robot 
interaction, and rehabilitation \cite{vinjamuri2009dimensionality, vinjamuri2014candidates}.

Several computational methods have been proposed for synergy extraction, including PCA \citep{d2006control, bizzi2008combining, muceli2010identifying}, and tensor factorization techniques \citep{ebied2019muscle, kim2020effect, chen2024muscle}. While these approaches have provided foundational insights, they typically assume linear mixing and do not naturally accommodate time-shifted, sparse, or temporally structured activation patterns. To address these limitations, time-varying and convolutional synergy models have been proposed \citep{d2005shared, chiovetto2022toward}. Our prior work has also showed the importance of modeling temporal alignment, activation delays, and dynamic structure in synergy extraction (see \citep{vinjamuri2009dimensionality, myrick2025robust}), showing that traditional linear factorization methods may miss critical components of the underlying sensorimotor organization. 


This work introduces a structured, scalable optimization framework for 
simultaneous synergy extraction and velocity reconstruction based on a 
sparse, time-shifted representation of grasping kinematics. Unlike existing two-stage methods that require separate data sets for the same task \citep{vinjamuri12}, our framework operates on a single data set.
Our main contributions are as follows:
\begin{itemize}
\item We develop a matrix-valued
representation of hand velocities in which synergies appear through time-shifted activations encoded by Toeplitz matrices. This representation captures temporal structure that traditional matrix-factorization methods do not model.

    \item We formulate a sparse optimization problem to jointly 
    identify a small set of synergies and task-dependent activation 
    coefficients. Our formulation uses group- and element-wise 
    sparsity, enabling automatic selection and pruning of inactive synergies.
    
    \item We develop an alternating minimization method (AMM) in which the coefficient
updates decouple across tasks, and each synergy is updated through a
separate Ridge regression. This structure enables the method to scale to large numbers of grasping tasks.

\end{itemize}

{\color{black}We validate the AMM on a natural-grasping dataset and an American
Sign Language posture dataset, and compare its performance against
the two-stage baseline. Our results show that AMM is an effective
dimensionality-reduction tool for synergy selection, while
reducing experimental data-collection requirements by roughly half.} 


\section{Preliminaries}\label{sec: prelims}
As said in the introduction, we assume that the latent synergy waveforms combine linearly to produce the observed hand movement. 
Let $v_i(t)\in \mathbb{R}$ denote the angular velocity of the $i$-th hand joint at time $t\in \mathbb{R}$. \cite{vinjamuri2009dimensionality} and the subsequent studies propose that 
\begin{equation}\label{vinjamuri-model}
    v^g_i(t)=\sum_{j=1}^{m}\sum_{k=1}^{K_j} c^g_{jk}s_{i}^{j}\left(t-t_{jk}\right), \quad i=1,\dots,n 
\end{equation}
for a grasping task $g\in \{1,\ldots,G\}$, where $s^j_i(t)$ corresponds to the $j$-th kinematic synergy associated with $i$-th hand movement. Here $K_j$ specifies how many times the $j$-th synergy is recruited, 
while $c^g_{jk}$ and $t_{jk}$ denote the nonnegative amplitude 
and time-shift parameters, respectively. 

\begin{figure}
  \centering
  \includegraphics[width=1.0\linewidth]{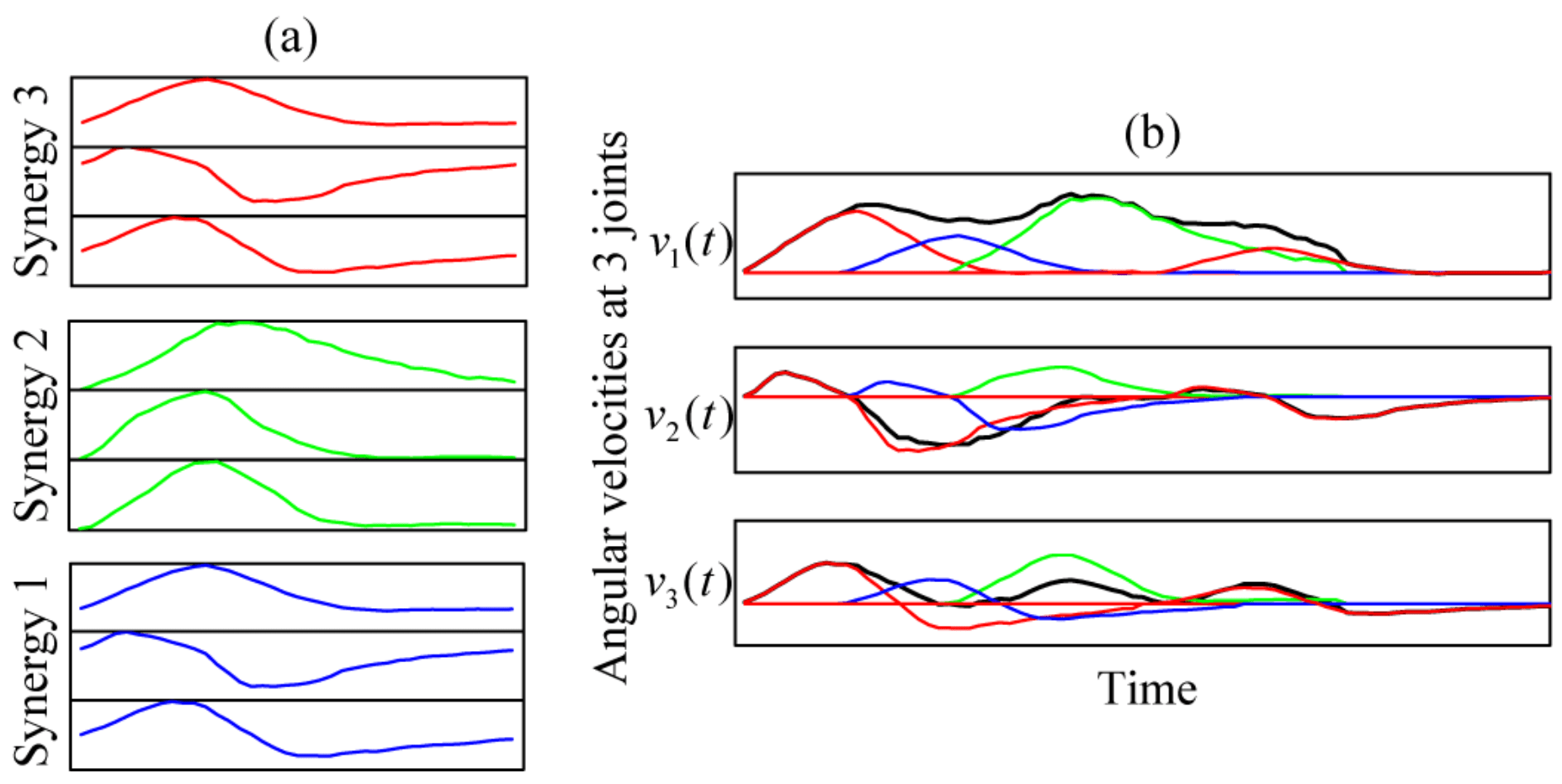}
  \caption{\footnotesize Three synergies---each repeated three times in (a)---combine to produce hand velocities (shown in black) at all joints in (b). Reproduced from \cite{vinjamuri2009dimensionality}.}
  \label{fig:motivation}
\end{figure}

The outer sum in \eqref{vinjamuri-model} indexes the $m$ synergies. The inner sum indexes for multiple recruitments (repeats) of each synergy.
Define the $n$-vector \( \mathbf{v}(t) = [v_1(t), \ldots, v_n(t)]^\top \) of angular velocities,  corresponding to the \( n \) hand joints at \( t \). Then, for all $n$-joints, from \eqref{vinjamuri-model} we have 
\begin{equation}\label{vinjamuri-model-vector}
    \mathbf{v}^{g}(t)=\sum_{j=1}^{m}\sum_{k=1}^{K_j} c^{g}_{jk}\mathbf{s}^{j}\left(t-t_{jk}\right).
\end{equation}
where 
each $n$-vector 
$\mathbf{s}^j(t) = [s_{1}^{j}(t), s_{2}^{j}(t), \dots, s_{n}^{j}(t)]^\top \in \mathbb{R}^n$. 
 Fig.~\ref{fig:motivation} visually displays the model in \eqref{vinjamuri-model-vector} for $n=3$ joints and $m=3$ synergies. In general, $n\ne m$. 




The governing equation in \eqref{vinjamuri-model-vector} admits a natural convolutive-mixture model interpretation. We outline the central idea here; see \cite{vinjamuri2008extraction, vinjamuri2009dimensionality} for details. Let $c_j(t)$ denote a train of impulses with amplitudes $c_{jk}$ occurring at times $t_{jk}$, where 
$k = 1, \ldots, K_j$. Then \eqref{vinjamuri-model-vector} becomes 
\begin{align}\label{eq: Conv}
    \mathbf{v}^g(t) = \sum_{j=1}^m(c^g_j * \mathbf{s}^j)(t).
\end{align}
Here $*$ is the standard convolution product for Linear Time Invariant signals.  

\setlength{\textfloatsep}{20pt plus 2.0pt minus 10.0pt}

\textbf{\textit{Problem Statement}}:
Given noisy measurements $\mathbf{v}^g(t)$ for $t = 1,\ldots,T$ and
$g = 1,\ldots,G$, identify the smallest number of synergies $m$ and the
fewest recruitments $K_j$ per synergy that jointly explain the observed
hand angular-velocity data across all $G$ tasks.

\section{A Matrix-valued Model}\label{sec: methods}
Since our goal is to learn synergy waveforms common to the \(G\) grasping tasks over a window of size \(T\), we stack the time-indexed $\mathbf{v}^g(t)$ in \eqref{vinjamuri-model-vector} to obtain a matrix-valued model of the activation coefficients and synergies. First, 
we express the time-shifted synergies using shift matrices by assuming
that the synergy duration satisfies $T_s \leq T$. This condition ensures that each synergy can be embedded into the observation window without truncation. 

Form the \(T_s\)-dimensional vector \(\mathbf{s}_{i}^{\,j} = [s_i^j(1), \ldots, s_i^j(T_s)]^\top\) by stacking over time the $i$-th element of the \(j\)-th synergy (see~\eqref{vinjamuri-model}). Define the template vector for the $j$-th synergy: 
\begin{align}\label{eq: synergy_vector}
\mathbf{s}_{[1:T_s]}^{\,j}
\triangleq 
\begin{bmatrix}
\mathbf{s}_{1}^{\,j}\\
\vdots\\
\mathbf{s}_{n}^{\,j}
\end{bmatrix}
\in \mathbb{R}^{nT_s}, 
\qquad
j \in \{1,\ldots,m\}.
\end{align}
Let $t_{jk}\geq 0$ be the $k$-th shift or recruitment associated with the $j$-th synergy. Define the $nT$-dimensional time-shifted vector $\mathbf{s}_{[1-t_{jk}:T-t_{jk}]}^{\,j}$ similar to \eqref{eq: synergy_vector}. Then there exists an $(nT) \times (nT_s)$ matrix $\mathbf{D}_{jk}$ of zeros and ones such that  
\begin{align}\label{eq: shift-mapping}
    \mathbf{s}_{[1-t_{jk}:T-t_{jk}]}^{\,j}= \mathbf{D}_{jk} \, \mathbf{s}_{[1:T_s]}^{\,j}. 
\end{align}
Henceforth, we drop the subscript notation from $\mathbf{s}_{[1:T_s]}^{\,j}$. In essence, \(\mathbf{D}_{jk}\) implements the time-shift by selecting and placing entries of \(\mathbf{s}^{\,j}\) into the correct time slots, with zeros filling the remaining positions (see Example \ref{ex: toy example}). 

So far our attention was on stacking time-indexed vectors. We focus on the spatial terms (that is, the number of synergies $m$ and their recruits). For the $j$-th synergy, define the matrix of time-shifted vectors given by \eqref{eq: shift-mapping}: 
\[
\mathbf{D}\!\left(\mathbf{s}^{\,j}\right)\triangleq \big[\, \mathbf{D}_{j1}\mathbf{s}^{\,j}\,|\, \mathbf{D}_{j2}\mathbf{s}^{\,j}\,|\, \dots\,|\, \mathbf{D}_{jK_{j}}\mathbf{s}^{\,j} \,\big]\in\real^{nT \times K_j}.
\]

\smallskip 
Let $\mathbf{c}_{j}^{g}=[c_{j1}^{g},
    \ldots, 
    c_{jK_j}^{g}]^\top$ be the vector of scalar coefficients multiplying the vectors $\mathbf{s}_i^j(t-t_{jk})$ in \eqref{vinjamuri-model-vector}. Then the matrices \(\mathbf{D}(\mathbf{s}^j)\) provide an efficient representation of the convolution model in \eqref{eq: Conv}. In particular, 
\begin{equation}\label{final-convolution-model}
\sum_{k=1}^{K_j} c^{g}_{jk}\mathbf{s}^{j}\left(t-t_{jk}\right)=
    \,\mathbf{D}\!\left(\mathbf{s}^{\,j}\right)\mathbf{c}_{j}^{g}. 
\end{equation}
As illustrated in Example~\ref{ex: toy example}, this matrix–vector product is a Toeplitz matrix acting on a coefficient vector, i.e., a convolution sum unfolded over the \(T\) time instants.

\medskip 

\begin{mdframed}[linewidth=0.8pt,
    leftmargin=3mm,
    rightmargin=3mm,
    innerleftmargin=4pt,
    innerrightmargin=4pt]
\begin{example}\label{ex: toy example}

We now present a toy model to visualize the matrices and vectors defined above. 
Our setup contains: \(G=4\) tasks, synergy duration of \(T_s=3\), and a movement duration of \(T=6\). Each synergy admits \(K_j = T - T_s + 1 = 4\) possible time shifts. For simplicity, the $k$-th time-shift value $t_{jk}=k$. 

Consider the case $n=m=1$. Let the shifting index $k=4$. 
The unshifted synergy segment ($nT_s=3$) is 
\[
\mathbf{s}^{\,j}_{[1:T_s]}
=
\begin{bmatrix}
\mathbf{s}^j(1)\\[2pt]
\mathbf{s}^j(2)\\[2pt]
\mathbf{s}^j(3)
\end{bmatrix}
\in \mathbb{R}^{3}.
\]
Note that \(j=1\); but we retain the notation \(j\). The time-shifted synergy vector of length \(nT=6\) is
\[
\mathbf{s}^{\,j}_{[1-t_{jk}:T-t_{jk}]}
=
\begin{bmatrix}
\mathbf{0}_{3}\\
\mathbf{s}^j(1)\\
\mathbf{s}^j(2)\\
\mathbf{s}^j(3)
\end{bmatrix}
\in \mathbb{R}^{6},
\]
where entries with indices outside \(\{1,\ldots,T_s\}\) are set to zero. This shift is given by the matrix: 
\[
\mathbf{D}_{j4}
=
\begin{bmatrix}
\mathbf{0}_{3\times 3}\\
\mathbf{I}_{3\times 3}
\end{bmatrix},
\qquad
\mathbf{s}^{\,j}_{[1-t_{jk}:T-t_{jk}]}
=
\mathbf{D}_{j4}\,
\mathbf{s}^{\,j}_{[1:T_s]}.
\]

Stacking the time-shifted versions of the synergy \(\mathbf{s}^{\,j}\) for all 
\(K_j = 4\) shifts yields a Toeplitz-like matrix: 
\begin{align*}
\mathbf{D}\!\left(\mathbf{s}^{\,j}\right)&=\big[\, \mathbf{D}_{j1}\mathbf{s}^{\,j}\,|\, \mathbf{D}_{j2}\mathbf{s}^{\,1}\,|\,
\mathbf{D}_{j3}\mathbf{s}^{\,j}\,|\,
\mathbf{D}_{j4}\mathbf{s}^{\,j}\,\big]\\
&=\left[
\begin{array}{c@{\hspace{14pt}}c@{\hspace{14pt}}c@{\hspace{14pt}}c}
\mathbf{s}^1(1) & 0        & 0        & 0       \\[6pt]
\mathbf{s}^1(2) & \mathbf{s}^1(1)   & 0        & 0       \\[6pt]
\mathbf{s}^1(3) & \mathbf{s}^1(2)   & \mathbf{s}^1(1)   & 0       \\[6pt]
0      & \mathbf{s}^1(3)   & \mathbf{s}^1(2)   & \mathbf{s}^1(1)  \\[6pt]
0      & 0        & \mathbf{s}^1(3)   & \mathbf{s}^1(2)  \\[6pt]
0      & 0        & 0        & \mathbf{s}^1(3)
\end{array}
\right]. 
\end{align*}

The velocity vector in \eqref{shift-matrix-convolution-model} becomes 
\begin{align*}
\begin{bmatrix}
    \mathbf{v}^g(1)\\
    \mathbf{v}^g(2)\\
    \vdots\\
    \mathbf{v}^g(6) 
\end{bmatrix}=\mathbf{D}\!\left(\mathbf{s}^{\,j}\right)\mathbf{c}_{j}^{g}=\begin{bmatrix}
    \mathbf{s}^1(1)c^g_{j1}\\
    \mathbf{s}^1(2)c^g_{j1}+\mathbf{s}^1(1)c_{j2}^{g}\\
    \vdots\\
\mathbf{s}^1(3)c^g_{jK}
\end{bmatrix}, 
\end{align*}
which is nothing but the convolution sum expressed for all $T=6$ instants. 

For \(n=2\) joint synergy vector of dimension \(nT_s = 6\), the shift matrix
is obtained by applying the same temporal synergy vector independently to each joint.
\end{example}
\end{mdframed}

For a grasping task $g\in \{1,\ldots,G\}$, define the time-stacked vector of velocities $\mathbf{v}^g_{1:T}=[(\mathbf{v}^g(1))^\top\ldots(\mathbf{v}^g(T))^\top]^\top$. From \eqref{vinjamuri-model-vector} and \eqref{final-convolution-model}, it then follows that
\begin{align}\label{shift-matrix-convolution-model}
    \mathbf{v}^g_{1:T}=
\,\sum_{j= 1}^m\mathbf{D}\!\left(\mathbf{s}^{\,j}\right)\mathbf{c}_{j}^{g}. 
\end{align}
Grasping velocity measurements are corrupted by additive sensor noise, so 
the model in \eqref{shift-matrix-convolution-model} can be augmented as
\begin{align}\label{eq: forward model}
    \mathbf{v}^g_{1:T}= \,\sum_{j=1}^m\mathbf{D}\!\left(\mathbf{s}^{\,j}\right)\mathbf{c}_{j}^{g}+\boldsymbol{\epsilon}^g.  
\end{align}
The set of matrices $\{\mathbf{D}(\mathbf{s}^j)\}_{j=1}^m$ is constant for all grasping tasks. We drop the subscript notation from $\mathbf{v}^g_{1:T}$.  Finally, define $\{\mathbf{s}^j\}\triangleq\{\mathbf{s}^1,\ldots,\mathbf{s}^m\}$; and $\{\mathbf{c}^g_j\}\triangleq\{\mathbf{c}^g_1,\ldots,\mathbf{c}^g_m\}$, for all grasping tasks $g\in \{1,\ldots,G\}$.


\section{Optimization Problem}

For the model in~\eqref{eq: forward model}, and using only the observed grasping velocities 
\(\mathbf{v}^1,\ldots,\mathbf{v}^G\), we seek to estimate:
\begin{enumerate}\itemsep4pt
    \item the synergy vectors \(\mathbf{s}^1,\ldots,\mathbf{s}^m\), under the assumption that only a small subset of these synergies is required to explain all grasping tasks;
    \item the coefficients \(\{\mathbf{c}_j^g\}\) for each task \(g\), assuming that each task activates only a few time-shifted synergies.
\end{enumerate}

These estimation goals can be formulated as an optimization problem equipped with regularizers (on the coefficient vectors $\mathbf{c}^g_1$). In particular, we solve
\begin{align}
\argmin_{\{\mathbf{s}^j\},\,\{\mathbf{c}_j^1\},\ldots,\{\mathbf{c}_j^G\}}
\; & 
\sum_{g=1}^G\left\lbrace 
\frac{1}{2} \bigl\lVert \mathbf{v}^{g} 
- \sum_{j=1}^m \mathbf{D}\!\left(\mathbf{s}^{\,j}\right)\mathbf{c}_{j}^{g} 
\bigr\rVert_2^2\right. \nonumber \\
&\qquad \; \left.+ \lambda\, \mathcal{R}\!\left(\{\mathbf{s}^{j}\}\right) 
+ \sum_{j=1}^m \|\mathbf{c}_j^g\|_{\ell2/\ell1}\right\rbrace,
\label{eq:main_opt}
\end{align}
where $\|\mathbf{x}\|_{\ell2/\ell1}\triangleq \lambda_1 \lVert \mathbf{x} \rVert_{2}+ \lambda_2 \lVert \mathbf{x} \rVert_{1}$ is the sparse group LASSO norm (norms are not squared); and \(\lambda, \lambda_1,\lambda_2 \ge 0\).


\looseness=-1 The first term in \eqref{eq:main_opt} is the squared $\ell_2$ loss.  
The second ensures that the learned synergies represent smooth patterns. We set $\mathcal{R}\!\left(\{\mathbf{s}^{j}\}\right)=\sum_j\|\mathbf{s}^j\|_2^2$. The last combines \(\ell_2\) and \(\ell_1\) norms to promote sparsity at two levels: only a few synergies are used (group sparsity), and each selected synergy is activate only a few time shifts (element-wise sparsity). 

If the problem in \eqref{eq:main_opt} is solved successfully, many
coefficient vectors \(\{\mathbf{c}^g_j\}\) become zero, yielding automatic
dimensionality reduction. The remaining nonzero vectors are sparse, so
the velocity in~\eqref{vinjamuri-model-vector} is captured by only a few
synergies and a small number of time-shifted activations. We present an alternating-minimization method for solving~\eqref{eq:main_opt}.

\vspace{-1.0mm}
\subsection{Alternating Minimization Method (AMM)}\label{subsec: altMin}

\looseness=-1 The problem in~\eqref{eq:main_opt} is non-convex and non-differentiable.  
The non-convexity arises from the bilinear terms \(\mathbf{D}(\mathbf{s}^j)\mathbf{c}_j^g\), where both factors are unknown.  
The non-differentiability stems from the \(\ell_2\) and \(\ell_1\) regularizers, although this issue is well-handled by modern optimization techniques. To address the non-convexity, we employ AMM, wherein we iteratively update $\{{\mathbf{c}}^{g}_j\}$ and $\{{\mathbf{s}}^j\}$ while holding the other fixed. 

AMM begins by randomly initializing $m=m_\text{int}$ synergies. If the AMM is converged and successful, we end up with fewer ($m_\text{final}\ll m_\text{int}$) synergies.

\textit{C-step (coefficient update):}  
Given the estimates of $\{\hat{\mathbf{s}}^j\}$, we update the coefficients by solving~\eqref{eq:main_opt} over \(\{\mathbf{c}^g_j\}\).  
Since \(\mathcal{R}(\cdot)\) acts only $\{\hat{\mathbf{s}}^j\}$, it is constant in this step and is dropped.  
The resulting optimization decouples across grasping tasks and can therefore be solved independently for each \(g\).
\begin{align}
\{\hat{\mathbf{c}}^{g}_j\}\leftarrow \arg\min_{\mathbf{\{c_j\}}} \frac{1}{2} \bigl\lVert \mathbf{v}^{g}
- \sum_{j=1}^m \mathbf{D}\!\left(\hat{\mathbf{s}}^j\right)\mathbf{c}_{j} 
\bigr\rVert_2^2+\sum_{j=1}^m||\mathbf{c}_{j}||_{\ell_2/\ell_1}. \label{eq: c-update}   
\end{align}
This is a convex optimization problem and can be solved efficiently \citep{simon2013sparse}. The penalty performs dimensionality reduction by driving entire sets of coefficients associated with a synergy to zero; 
see Remark \ref{rmk}.

\textit{S-step (synergies update):} Given $\{\hat{\mathbf{c}}^{g}_j\}$, for $g\in\{1,\ldots,G\}$, we solve the problem below after dropping the constant term (for this step): 
\begin{align}\label{eq: S-update}
\{\hat{\mathbf{s}}^j\}\leftarrow\argmin_{\{\mathbf{s}^j\}}
\; & 
\sum_{g=1}^G
\frac{1}{2} \bigl\lVert \mathbf{v}^{g} 
- \sum_{j=1}^m \mathbf{D}\!\left(\mathbf{s}^{\,j}\right){\hat{\mathbf{c}}}_{j}^{g} 
\bigr\rVert_2^2 + \frac{\alpha}{2}\sum_{j=1}^m\|\mathbf{s}^j\|_2^2,
\end{align}
where we set $\lambda=\alpha/2G\geq 0$. 

\looseness=-1 At the outset, solving~\eqref{eq: S-update} appears challenging.  
First, \(\mathbf{s}^j\) influence every task, so the $l_2$ loss couples all \(G\) summands, which could be large.  
Second, each \(\mathbf{D}(\mathbf{s}^j)\) is an \(nT \times K_j\) Toeplitz matrix whose size grows with the number of shifts \(K_j\), making a naive implementation expensive. After some algebra, we show that the problem in~\eqref{eq: S-update} can be solved efficiently, one synergy \(\mathbf{s}^j\) at a time, by reducing it to a simple \(\ell_2\)-regularized least-squares (i.e., a Ridge regression).

For any \(j \in \{1,\ldots,m\}\), define the residual for the task \(g\) excluding the \(j\)-th synergy:
\[
\mathbf{r}_{-j}^{g}
= 
\mathbf{v}^{g}
- \sum_{\ell \ne j} \mathbf{D}(\mathbf{s}^{\ell}) \hat{\mathbf{c}}_{\ell}^{g},
\qquad g=1,\ldots,G.
\]
By direct inspection, for any fixed synergy vector \(\mathbf{s}^j\),
\[
\mathbf{D}\!\left(\mathbf{s}^{\,j}\right)\hat{\mathbf{c}}_{j}^{g}
=
\left(\sum_{k=1}^{K_j} \hat{c}_{jk}^g \mathbf{D}_{jk}\right)\mathbf{s}^j,
\qquad g=1,\ldots,G.
\]
Let \(\mathbf{B}_{j}(\hat{\mathbf{c}}_{j}^{g})\) denote the matrix in parentheses to the right, which is known because \(\hat{\mathbf{c}}_{j}^{g}\) is known.

Stack the quantities across the \(G\) tasks:
\[
\mathbf{r}_{-j}
=
\begin{bmatrix}
\mathbf{r}_{-j}^{1} \\
\vdots \\
\mathbf{r}_{-j}^{G}
\end{bmatrix}
\in \mathbb{R}^{nTG};
\,\,\,
\mathbf{B}_{j}
=
\begin{bmatrix}
\mathbf{B}_{j}(\hat{\mathbf{c}}_{j}^{1}) \\
\vdots \\
\mathbf{B}_{j}(\hat{\mathbf{c}}_{j}^{G})
\end{bmatrix}
\in \mathbb{R}^{(nTG)\times (nT_s)}.
\]
The update for the \(j\)-th synergy reduces to Ridge regression:
\[
\hat{\mathbf{s}}^{\,j}
\leftarrow
\arg\min_{\tilde{\mathbf{s}}}
\frac{1}{2}
\lVert \mathbf{r}_{-j} - \mathbf{B}_{j} \tilde{\mathbf{s}} \rVert_2^{2}
+
\frac{\alpha}{2}
\lVert \tilde{\mathbf{s}} \rVert_2^{2},
\]
where \(\tilde{\mathbf{s}}\) has the same dimension as \(\mathbf{s}^{\,j}\).

After each update, we normalize \(\mathbf{s}^{\,j}\) and rescale \(\mathbf{c}_{j}^{g}\) accordingly to remove scaling ambiguity.

\medskip 
\begin{remark}\label{rmk}
{\color{black} For synergy extraction, we may use only the group LASSO regularizer
(i.e., set $\lambda_2 = 0$ in \eqref{eq:main_opt}) instead of the full sparse-group penalty. 
Nevertheless, we found that using the sparse group lasso during training improves generalization during testing. Specifically, it activates only a few shifts for the active synergies. \QEDW }
\end{remark}

\section{Results and Discussion}
\looseness=-1 We evaluate the performance of the AMM on hand grasping data and compare it against
the two-stage approach of \cite{vinjamuri2009dimensionality}. Our
simulations use the natural-grasp dataset, in which subjects\footnote{We consider a total of seven subjects in this work.} grasp objects
of various shapes and sizes while wearing a right-handed CyberGlove that
records joint-angle measurements. We focus on ten joints: the MCP and IP
joints of the thumb and the MCP and PIP joints of the four fingers, which
capture the dominant kinematic patterns in the grasping tasks. 

AMM uses only the natural-grasp dataset for both synergy extraction and
coefficient estimation. In contrast, the two-stage baseline requires an
additional rapid-grasp dataset to extract synergies (via SVD) before estimating
coefficients from natural grasps, doubling the
data-collection effort. Datasets were generated through real-time
experiments conducted by one of the authors in prior work
\citep{vinjamuri2009dimensionality, vinjamuri12}.


\subsection{Synergy Extraction and Coefficient Estimation}\label{subsec: training}

\vspace{-2.0mm}

Each grasp consists of \(T = 82\) time samples across
\(n = 10\) joints, yielding \(G = 100\) grasping tasks. We
set \(m_{\text{int}} = 10\) candidate synergies, each of duration
\(T_s = 39\) samples. These synergies are then shifted and embedded into
the observation window as described in Section~\ref{sec: methods}. This
gives each synergy \(K_j = T - T_s + 1 = 44\) possible time shifts. Thus, for
ten joints, the coefficient vector associated with the \(j\)-th synergy in
task \(g\) is \(440\)-dimensional. 
{\color{black} The tuning parameters ($\lambda, \lambda_1,\lambda_2$ in \eqref{eq:main_opt}) for the sparse group LASSO and the ridge regression were selected using a small grid search over a discrete set of reasonable values.}

\begin{figure}
    \centering
    \includegraphics[width=1.0\linewidth]{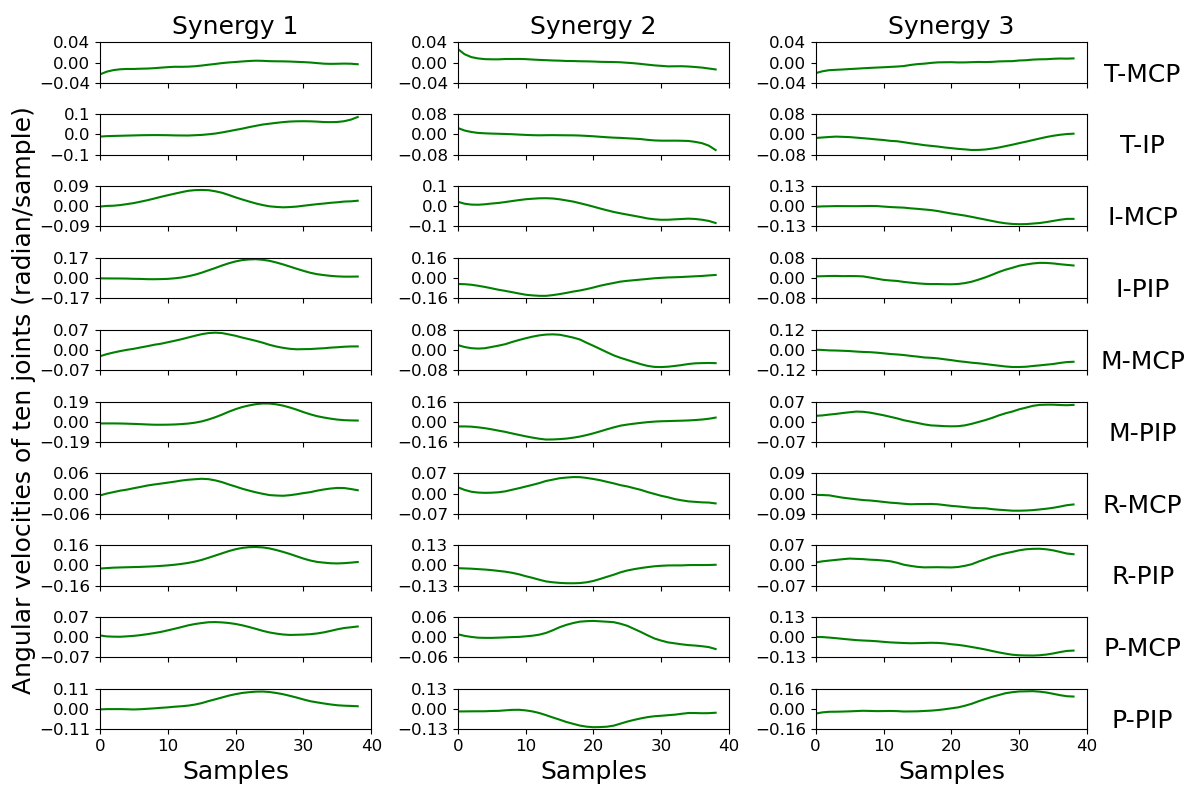}
    \caption{\small Joint angular-velocity profiles of three synergies among the seven extracted using AMM. The smoothness of the profiles are due to the smoothing penalty term $\mathcal{R}(\cdot)$ in \eqref{eq:main_opt}. Abbreviations: T, thumb; I, index finger; M, middle finger; R, ring finger; P, pinky finger; MCP, metacarpophalangeal
joint; IP, interphalangeal joint; PIP, proximal IP joint.}
    \label{fig:synergy_graphs}
\end{figure}

AMM discarded three synergies that remained inactive across all 100
grasping tasks, yielding \(m_{\text{final}} = 7\) synergies. The joint
angular-velocity profiles of three of these synergies for Subject~2 are in Fig.~\ref{fig:synergy_graphs}. Each row corresponds to one of the
ten joints. As expected, the synergy waveforms are smooth functions of time with peaks and valleys marking periods of hand
acceleration and deceleration.

\setlength{\textfloatsep}{5pt plus 2.0pt minus 10.0pt}

\begin{figure*}[t]
    \centering
    \includegraphics[width=1.0\linewidth]{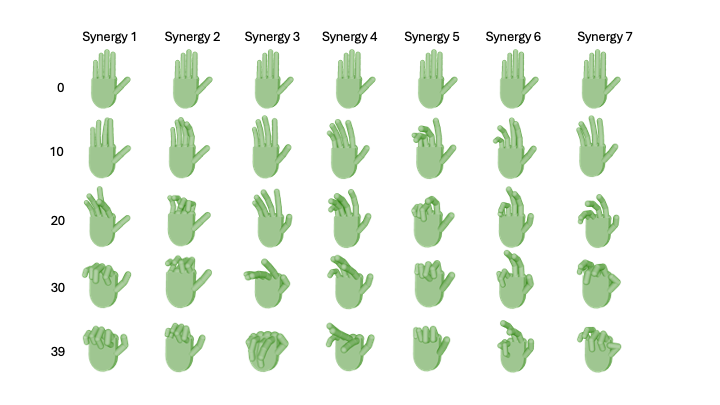}
    \caption{\footnotesize \color{black} Joint angle profiles of the seven synergies at time steps 0, 10, 20, 30, and 39.  Columns represents synergies and rows represent the time steps. Each hand posture is produced by coordination of ten finger joints. For instance, in Synergy 7, the MCP and IP joints of the thumb and the MCP and PIP joints of the other four fingers combine to form the common two-finger pinch posture.}
    \label{fig:3-d_synergies}
\end{figure*}

To visualize each synergy’s temporal postures (that is, the corresponding
hand movements), we recover joint angles by discretely integrating the joint
angular velocities. The seven resulting postural synergies are displayed in
Fig.~\ref{fig:3-d_synergies}. Each synergy is illustrated using five
snapshots at 0\%, 25\%, 50\%, 75\%, and 100\% of the task duration,
revealing how the hand configuration evolves over the course of the
movement. The end posture is particularly informative, as it reflects the specific
hand configuration that each synergy contributes toward achieving a given
grasp.




\setlength{\textfloatsep}{10pt plus 2.0pt minus 10.0pt} 

We compare the (end) postural synergies obtained using AMM with those
reported by the two-stage baseline in \cite{vinjamuri2009dimensionality}. {Although the baseline estimates synergies from the rapid-grasp dataset,
the comparison is valid because both rapid and natural grasps involve the
same set of grasping tasks.}
 In Fig.~\ref{fig:3-d_synergies} and \cite[Fig. 6a]{vinjamuri2009dimensionality}, 
  we observe three dominant terminal postures: (1) a whole-hand (power) grasp, characterized by simultaneous flexion of all digits; (2) a two-finger pinch, primarily involving coordinated flexion of the thumb and index finger; and (3) a full-hand extension posture in which all digits return to an extended configuration. 
  
\looseness=-1  The emergence of these same temporal postures across studies -- despite differences in synergy extraction methods -- highlights a fundamental underlying structure in human hand control. Importantly, the fact that these archetypal postures arise from both standard SVD \citep{vinjamuri2009dimensionality} and our joint estimation problem in \eqref{eq:main_opt} (current work) demonstrates that each method converges on a set of anatomically and functionally meaningful synergies. 

\vspace{-1.0mm}

\subsection{Reconstruction of Hand Movements}\label{subsec: testing}
By combining synergy
velocities (shown in Fig.~\ref{fig:synergy_graphs}) with their estimated time-shifted coefficients (not shown) as in
model \eqref{vinjamuri-model}, we obtain reconstructed joint velocities.
The difference between these reconstructions and the measured velocities
yields the training error. While important, we
shall not focus on training error as our interest is in
how well the learned synergies generalize to unseen grasps. We therefore
turn to testing-phase reconstruction.

In the testing phase, we use the American Sign Language (ASL) dataset
described in \cite{vinjamuri2009dimensionality}. This dataset consists of 36 ASL gestures (the digits 0-9 and the letters A-Z), collected using the same Cyber-Glove and the same ten selected joints as in training. This choice of the data serves two purposes:
(i) it broadens the scope of our evaluation, and (ii) it highlights the
core idea behind synergies: they act as basic building
blocks of movement. Once learned from one set of tasks (e.g., natural
grasping), the same synergies can be used to reconstruct entirely
different types of hand motions, such as ASL gestures.


Accordingly, for both our method and the two-stage baseline, we use
the synergies learned during training. The activation coefficients from
training are not reused; in the testing phase, these coefficients are the
unknowns and are determined by solving the $\ell_1$-regularized
least-squares:
\begin{align}\label{eq: LASSO}
\hat{\mathbf{c}}^{\,g}
\;\leftarrow\;
\argmin_{\mathbf{c}^g}
\;\frac{1}{2}\Bigl\|\mathbf{v}^g_{\text{test}}
- \mathbf{B}^\top\,\mathbf{c}^g\Bigr\|_2^2
+ \lambda_{\text{test}}\|\mathbf{c}^g\|_1,
\end{align}
for each testing gesture\footnote{In the ASL dataset, each movement is termed a gesture rather than
a grasping task, but we retain the notation $g$ for indexing.} $g = 1,\ldots,36$, with $\lambda_{\text{test}} \ge 0$.
Here $\mathbf{v}^g_{\text{test}} \in \mathbb{R}^{nT}$ is the measured
angular-velocity trajectory for the testing data, obtained by concatenating
the $n=10$ joint velocities over the testing duration ($T = 86$ samples). The
matrix $\mathbf{B}$ is the bank of time-shifted synergies arranged in the
canonical order described in \cite{vinjamuri12}. Under an appropriate
permutation, one can verify that $\mathbf{B}\mathbf{c}^g$ recovers the
linear model in \eqref{shift-matrix-convolution-model} (see Remark \ref{rmk2}).

Once $\hat{\mathbf{c}}^{\,g}$ is obtained, the model in
\eqref{vinjamuri-model} is used to reconstruct the joint velocities,
which are then compared with the measured testing velocities to compute
the reconstruction error. A smaller error indicates that the mixture model
captures the underlying hand movements and that the estimation procedure
(AMM or the two-stage baseline) successfully identifies its parameters.

The (normalized) reconstruction error for each ASL gesture $g$ is defined
as in \cite{vinjamuri2009dimensionality}:
\[
\frac{\sum_{i=1}^{n}\sum_{t=1}^{T}\bigl[v_i^g(t)-\hat{v}_i^g(t)\bigr]^2}
     {\sum_{i=1}^{n}\sum_{t=1}^{T}v_i^g(t)^2},
\]
where $v_i^g(t)$ denotes the measured angular velocity of joint $i$ at time
$t$, and $\hat{v}_i^g(t)$ is its reconstructed counterpart. For the
36 ASL gestures ($g=1,\ldots,36$), the per-subject average reconstruction
errors are reported in Table~\ref{tab:errors}. Across all seven subjects,
the global average reconstruction error is $0.2783$ with a standard
deviation (std) of $0.02153$. For a fixed set of seven synergies, this average
error is comparable to that reported for the baseline two-stage method
\citep[Fig.~8b]{vinjamuri2009dimensionality}.

\begin{table}[h]
\centering
\caption{\footnotesize Average reconstruction errors of 36 ASL gestures for each subject.}
\begin{tabular}{|c|c|c|c|}
\hline 
Subject 1 & Subject 2 & Subject 3 & Subject 4 \\ \hline
0.2587    & 0.2799    & 0.2803    & 0.2486    \\ \hline
Subject 5 & Subject 6 & Subject 7 & Avg error (std)       \\ \hline
0.2976    & 0.3113    & 0.2715    & \textbf{0.2783 (0.02153) }  \\
\hline 
\end{tabular}
\label{tab:errors}
\end{table}

\begin{figure}[t]
  \centering
  \includegraphics[width=1.0\linewidth]{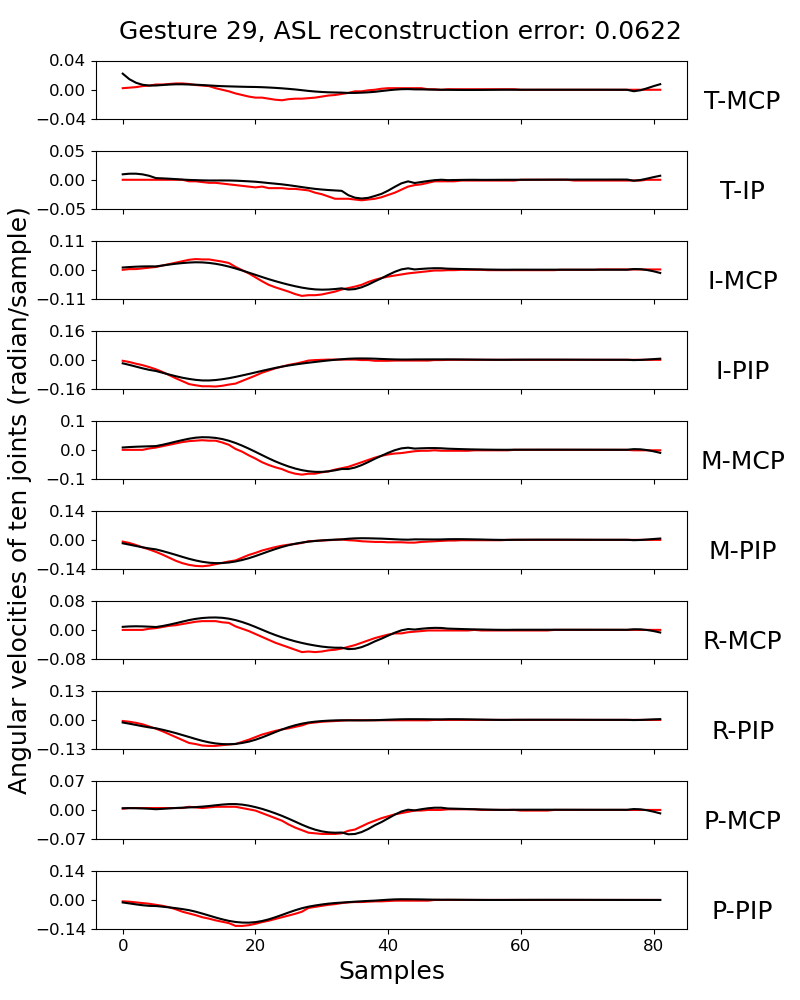}
  \caption{\footnotesize Joint angular velocity of an ASL gesture (red line) and our model's reconstruction (black line) using seven synergies. See Fig.~\ref{fig:synergy_graphs} for abbreviations of finger joints.}
  \label{fig:asl_29_recon_graph}
\end{figure}
\begin{figure}[t]
    \centering
    \includegraphics[width=1.0\linewidth]{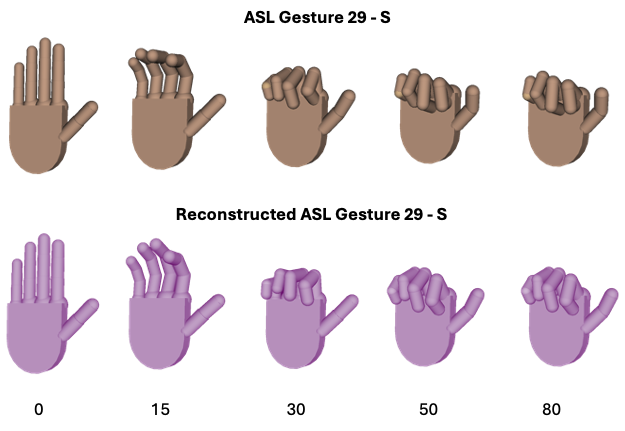}
    \caption{\footnotesize Joint angle profiles of ASL gesture 29 (the letter \textit{S}) and our model's reconstruction at time steps 0, 15, 30, 50, and 80.}
    \label{fig:asl_29_recon_model}
\end{figure}

An example of a successful reconstruction for a particular ASL movement is depicted in
Fig.~\ref{fig:asl_29_recon_graph}, with the corresponding temporal
postures at selected time points in Fig.~\ref{fig:asl_29_recon_model}.
Both figures use synergies extracted from Subject 2’s natural-grasping
data to reconstruct the ASL gesture sequence from the same subject.

\medskip 





\setlength{\textfloatsep}{10pt plus 2.0pt minus 10.0pt}

\setlength{\textfloatsep}{10pt plus 2.0pt minus 10.0pt}

\begin{remark}\label{rmk2} \textit{(Implementation Details)}
{\color{black} Before solving the velocity reconstruction problem in \eqref{eq: LASSO}, we
mitigate multicollinearity in the synergy bank matrix $\mathbf{B}^\top$ by
removing shifts whose pairwise correlation exceeds a threshold $\tau=0.8$.
This reduces multicollinearity in $\mathbf{B}^\top$, which otherwise can lead
to inconsistent coefficient estimates during testing.
To ensure that the reconstructed velocities remain smooth functions of
time, we apply a Savitzky–Golay filter to the columns of the resulting
matrix $\mathbf{B}^\top$. \QEDW
}



\end{remark}


\section{Conclusion} 

{\color{black} This work introduced a group-sparsity–based optimization framework for
learning time-shifted hand synergies and reconstructing joint velocities
from kinematic data. Central to this framework is a matrix-valued
nonlinear equation that links the observed data to the synergy waveforms
and their time-shifted activation coefficients. We developed an
iterative method that updates activation coefficients
across tasks and refines synergy waveforms through independent Ridge
regressions. Experiments on natural-grasping and ASL posture datasets demonstrate that our method extracts compact and
interpretable synergies while achieving accurate velocity reconstruction.

The numerical results are encouraging and point to
several directions for future work: developing convergence
guarantees for AMM, exploring structure-promoting
regularization schemes, and extending the framework to multimodal or
higher-dimensional datasets. Improving computational scalability may also
enable applications in sports performance, rehabilitation, and motor-skill
training.
}


\bibliography{ifacconf}             
                                                   







\appendix
\end{document}